# Features Reconstruction Disentanglement Cloth-Changing Person Re-Identification

Zhihao Chen[1][0009-0000-6806-1326], Yiyuan Ge[2][0009-0006-5442-1865], Qing Yue[1*][0009-0003-8145-7394]

[1] the School of Computer, Beijing Information Science and Technology University, Beijing, China
[2] the School of Instrument Science and Opto-Electronics Engineering, Beijing Information Science and Technology University, Beijing, China
2021011561@bistu.edu.cn, geyiyuan@bistu.edu.cn, yueqing99@sohu.com

**Abstract.** Cloth-changing person re-identification (CC-ReID) aims to retrieve specific pedestrians in a cloth-changing scenario. Its main challenge is to disentangle the clothing-related and clothing-unrelated features. Most existing approaches force the model to learn clothing-unrelated features by changing the color of the clothes. However, due to the lack of ground truth, these methods inevitably introduce noise, which destroys the discriminative features and leads to an uncontrollable disentanglement process. In this paper, we propose a new person re-identification network called features reconstruction disentanglement ReID (FRD-ReID), which can controllably decouple the clothing-unrelated and clothing-related features. Specifically, we first introduce the human parsing mask as the ground truth of the reconstruction process. At the same time, we propose the far away attention (FAA) mechanism and the person contour attention (PCA) mechanism for clothing-unrelated features and pedestrian contour features to improve the feature reconstruction efficiency. In the testing phase, we directly discard the clothing-related features for inference, which leads to a controllable disentanglement process. We conducted extensive experiments on the PRCC, LTCC, and Vc-Clothes datasets and demonstrated that our method outperforms existing state-of-the-art methods.

**Keywords:** Person Re-identification, Cloth-changing Person Re-Identification, Feature Reconstruction Disentanglement.

## 1      Introduction

The task of person re-identification is to match pedestrians across scenes and viewpoints [1-4][64-66], which assumes that the pedestrian's clothing does not change. However, when considering matching pedestrians over long time spans, the clothing of pedestrians changes, which leads to the failure of existing person re-identification systems. To address the above problem, researchers have proposed the cloth-changing person re-identification (CC-ReID) task, which aims to match pedestrians across clothing.

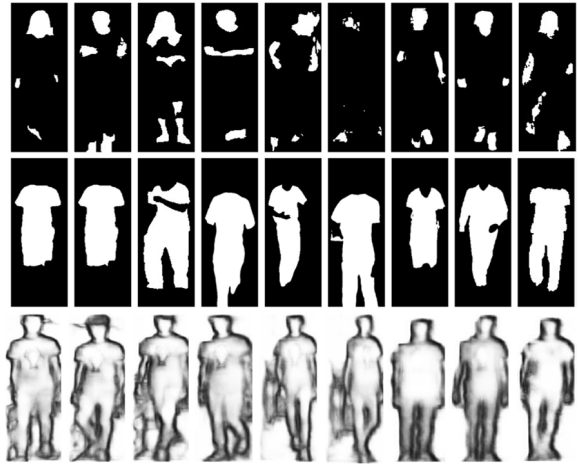

**Fig. 1.** The upper figure shows the human parsing mask, the first row shows the clothing urelated features map, the second row shows the clothing-related features maps, and the third row shows the pedestrian contour map.

Currently, there are two categories of mainstream CC-ReID methods: fusion-based methods [21] [49] and disentanglement-based methods [34] [50]. Typically, disentanglement-based methods are superior to fusion-based methods [13]. Fusion-based methods cannot fully enumerate and fuse all clothing-unrelated features. In contrast, disentanglement-based methods avoid this problem by separating these features from the original image. There are two main categories of existing disentanglement-based methods: picture-generating methods [21][51-53] and multimodal methods [16][54-55]. In order to reduce the interference of clothing on re-identification, picture-generating methods force the model to learn clothing-unrelated features by randomly changing the pixel points in the clothing region, which solves the problem that the traditional method fails in cloth-changing scenarios. However, picture-generating methods are also limited by some negative factors [21], e.g., the most discriminative clothing-unrelated features such as face, legs, and body shape may be corrupted when generating pictures. Instead, multimodal methods avoid feature corruption by introducing additional branches to learn clothing-unrelated features [16][53]. [6][54-55] hope that the features extracted from RGB images will be close to those extracted from other modalities, such as greyscale images, contour images, etc., to overcome the effect of changes in clothing color on feature extraction. However, this approach is still flawed as it may ignore color-related but discriminative clothing-unrelated features such as skin, bag, and shoe color.

Introducing feature reconstruction [57-58] mitigates the limitations of the CC-ReID method mentioned above. [6][16-18][20-22] achieve disentanglement by restructuring clothing-unrelated and clothing-related features of pedestrians and mapping them to the RGB domain. However, the lack of ground truth prevents direct supervision and control of the reconstruction process. These methods can only be supervised indirectly using additional discriminators [6][16-17], regularisation terms [6][18][20-22][67-68], or mapping back to the source image [17-18][23]. This results in clothing-unrelated features inevitably interfered with by clothing-related features, limiting the reconstruction results and model performance.

If the reconstruction process becomes controllable, then the accuracy of CC-ReID can be improved [6][18][69]. This requires one-to-one correspondence of the extracted clothing-unrelated features, clothing-related features with the clothing regions, and non-clothing regions in the pedestrian image. We introduce a human parsing mask to guide the reconstruction of clothing-unrelated features, clothing-related features, and pedestrian contour features, which controllably decouples clothing-unrelated and clothing-related information in the feature space and, thus, more efficiently utilizes clothing-unrelated features for re-identification. As shown in Fig. 1, the clothing-unrelated features map contains only local features such as head, hands, shoes, etc., scattered and widely spaced. For this reason, we design the far away attention (FAA) mechanism for global modeling. Pedestrian contour maps are usually blurry and dim, so we design the person contour attention (PCA) mechanism to aggregate different levels of edge information and enhance the pedestrian contour information. Finally, in the inference phase, we achieve controlled disentanglement using only clothing-unrelated features for re-identification. The main contributions of this paper are as follows:

- We propose a novel clothing-changing person re-identification network features reconstruction disentanglement ReID (FRD-ReID), which achieves controlled decoupling of clothing-unrelated and clothing-related features by reconstructing deep features extracted by the backbone.
- For reconstructing clothing-unrelated features, we propose the far away attention (FAA) mechanism for the global modeling of sparsely distributed features.
- To reconstruct pedestrian contour features, we design the person contour attention (PCA) mechanism to aggregate and enhance the different levels of edge information to make the pedestrian contour clearer.

## 2    Related Works

### 2.1    Person Re-Identification

Person re-identification identifies and matches target pedestrians from an existing video or image sequence with the assumption that people do not change their clothes for a short period. Therefore, we can use clothes' color and texture information to extract discriminative features for person re-identification. Mainstream methods use CNN architectures in the following three categories: 1. Representing pedestrians by global features, Luo et al. [11] proposed several methods to improve the recognition of global features. However, the global feature-based approach performs poorly when faced with occlusion and pedestrian posture changes. Zheng et al. [60] treated each individual as a separate category and used a multi-category loss function to improve the discriminative properties of global features. However, global features cannot distinguish fine differences, such as hair, shoes, and so on. Therefore, the method of using local discriminative features has been proposed. 2. Using local features, dividing the features extracted from the backbone into several parts captures pedestrians' locally salient information. Sun et al. [10] used visibility-aware part-level features to solve the problem of incomplete images and spatial mismatch to some extent. Zheng et al. [25] vertically decomposed deep convolutional features into multiple parts and then used each part to learn independently to improve local features. 3. Combining global and local features is used to obtain a more significant representation of pedestrian features. Wang et al. [12] proposed Multigranular Network (MGN) in order to combine fine-grained local features with global features.

However, when we consider a longer period, people change their clothes irregularly, which makes it an interfering feature, leading to the poor performance of the above methods. Therefore, cloth-changing person re-identification (CC-ReID) has received increasing attention recently.

### 2.2   Feature Reconstruction in Cloth-Changing Person Re-Identification

Unlike person re-identification, cloth-changing person re-identification (CC-ReID) requires capturing clothing-unrelated features. Feature reconstruction is mainly used to decouple clothing-related features from clothing-unrelated features in CC-ReID. Xu et al. [16] enabled the model to better learn clothing-unrelated features by reconstructing exemplars within the same class to reduce the variation of features and simultaneously generate adversarial dressing exemplars between different classes. Li et al. [17] use contour and texture information from greyscale images and combine it with different dresses of the same person to restore the original image. Zheng et al. [59] proposed an end-to-end joint learning framework to optimize the feature reconstruction process. It decomposes each instance into appearance and structure codes, then combines the codes of different instances to generate reconstructed images and feeds them to the generator and the discriminator. Due to the lack of ground truth, the methods above cannot be directly supervised and controlled. Therefore, the extracted features do not guide the model learning better.

## 3   Methods

### 3.1   Architecture

As shown in Fig. 2, our network consists of three branches: the clothing classifier, the ID classifier, and the human reconstruction branch (HRB). In this section, we first introduce the overall architecture of feature reconstruction disentanglement ReID (FRD-ReID). Subsequently, we will illustrate the human reconstruction branch (HRB) and its internal components in detail.

### 3.2   Human Reconstruction Branch

Clothing-unrelated and clothing-related features correspond to the non-clothing and clothing region of the pedestrian, respectively, and the disentanglement of clothing-unrelated and clothing-related features is achieved by controllably reconstructing these regions. We note the close relationship between the clothing-unrelated and clothing-related features of the pedestrian contour, so the feature reconstruction is developed in three parts. Suppose the feature vector extracted by backbone is $x \in R^{H \times W \times C}$:

$$F_{uncloth}, F_{cloth}, F_{contour} = Split(x) \quad (1)$$

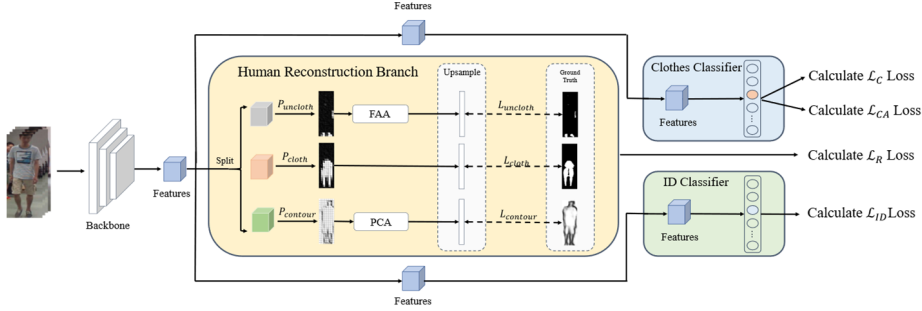

**Fig. 2.** The overall architecture of features reconstruction disentanglement ReID(FRD-ReID). It consists of three branches, which are human reconstruction branch, clothes classifier, and ID classifier.

Where $Split(\cdot)$ denotes splitting in the channel dimension, and $F_{uncloth} \in R^{H \times W \times \frac{C}{4}}$, $F_{cloth} \in R^{H \times W \times \frac{C}{2}}$, and $F_{contour} \in R^{H \times W \times \frac{C}{4}}$ denote clothing-unrelated, clothing-relevant, and pedestrian contour features, respectively. We split and reconstruct the feature maps extracted from the backbone network in the channel dimension, which avoids the information fusion between channels and the interference between different features and provides a guarantee for controllable disentanglement.

As shown in Fig. 1, the clothing-unrelated features only contain local features such as head, hand, shoes, etc. These features are scattered and far away from each other, so the traditional attention mechanism can't pay attention to them well. We design the far away attention mechanism, which can model clothing-unrelated features globally.

Through Fig. 1, we can observe that the pedestrian contour is coherent, and we hope to enhance the positional correlation and strengthen the extraction ability simultaneously. To this end, we designed the person contour attention (PCA) mechanism, which aggregates and enhances the edge information at different levels to improve the contrast between the pedestrian contour and the background and make the contour clearer.

**Far Away Attention Mechanism.** The far away attention (FAA) mechanism directly captures remote dependencies by computing the interaction between two locations. $F_{uncloth} \in R^{H \times W \times \frac{C}{4}}$ is the clothing-unrelated features map, and we use three $1 \times 1$ convolutional processes to obtain three features maps, denoted as $F_q$, $F_k$, and $F_v$.

$$F_q = Conv_1(F_{uncloth}, \theta) \quad (1)$$
$$F_k = Conv_1(F_{uncloth}, \theta) \quad (2)$$
$$F_v = Conv_1(F_{uncloth}, \theta) \quad (3)$$

$Conv_1(\cdot, \theta)$ represents $1 \times 1$ convolution and $\theta$ is the process parameter. Then $F_q$ and $F_k$ are matrix multiplied followed by using Softmax function to get $F_{qk}$. The above process can be formulated as:

$$F_{qk} = Softmax(F_q \otimes F_k) \quad (4)$$

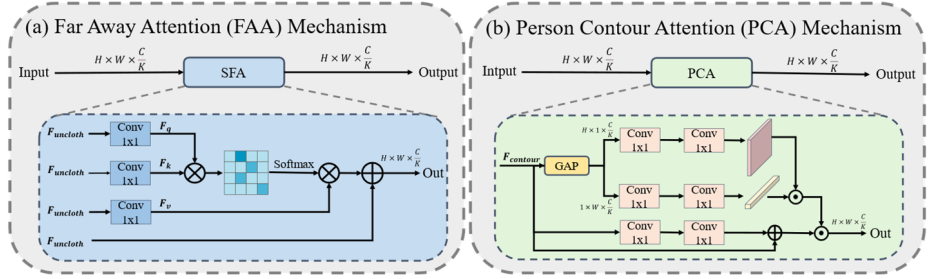

**Fig. 3.** The Fig.3(a) is the detail architecture diagram of far away attention (FAA) mechanism. The Fig. 3(b) is the diagram of person contour attention (PCA) mechanism.

$Softmax(\cdot)$ is the softmax activation function and $\otimes$ represents pixel-level multiplication. We then matrix multiply $F_{qk}$ with $F_v$ to obtain $F_{qkv}$. To ensure stability during training, we also use the residual strategy:

$$F_{qkv} = F_{qk} \otimes F_v \qquad (5)$$
$$F_{faa} = F_{qkv} \oplus F_{uncloth} \qquad (6)$$

Where $\otimes$ denotes pixel level multiplication, $\oplus$ denotes residual link, and $F_{faa}$ is the result obtained after residual link.

**Person Contour Attention Mechanism.** As shown in Fig. 1, the pedestrian contour map is usually fuzzy and has low contrast, making it difficult to extract clear and complete information. Therefore, we design the person contour attention (PCA) mechanism to aggregate and enhance different levels of edge information, improving the contrast between the pedestrian contour and the background to make the contour clearer. Given a pedestrian contour map $F_{contour} \in R^{H \times W \times \frac{C}{4}}$, firstly, global average pooling is performed on $F_{contour}$ to obtain the global representation, and the feature maps with length and width (1,w) and (h,1) are obtained, which are denoted as $F_h \in R^{1 \times W \times \frac{C}{4}}$ and $F_w \in R^{H \times 1 \times \frac{C}{4}}$, respectively. The above process can be expressed as:

$$F_h, F_w = GAP(F_{contour}) \qquad (7)$$

$GAP(\cdot)$ denotes the global average pooling function. Then, we use $1 \times 1$ convolution to increase the dimension of the feature map to $\frac{2C}{4}$ and then use the same size of convolution kernel to reduce the dimension to $\frac{C}{4}$. The above process can be expressed as follows:

$$F'_h = Conv_1(Conv_1(F_h, \theta), \theta) \qquad (8)$$
$$F'_w = Conv_1(Conv_1(F_w, \theta), \theta) \qquad (9)$$

Where $Conv_1(\cdot, \theta)$ denotes $1 \times 1$ convolution, $\theta$ is the process parameter, $F'_h$ and $F'_w$ are the feature maps obtained after convolution processing. Next, $F'_h$ is matrix dot-multiplied with $F'_w$ to obtain a feature map of size $H \times W \times \frac{C}{4}$. The above process can be expressed as:

$$F'_{hw} = F'_h \odot F'_w \qquad (10)$$

Where $\odot$ denotes the matrix dot product and $F'_{hw}$ is the result after dot product. In the person contour attention mechanism, we also introduce residual branching. The pedestrian contour map $F_{contour}$ is firstly upgraded and then downgraded using $1 \times 1$ convolution to obtain $F'_{contour}$, and then it is residually connected with the contour map $F_{contour}$. The above process can be expressed as:

$$F'_{contour} = Conv_1(Conv_1(F_{contour}, \theta), \theta) \tag{11}$$

$$F_{res}^{Contour} = F'_{contour} \oplus F_{contour} \tag{12}$$

Where $\oplus$ denotes residual link, $Conv_1(\cdot, \theta)$ denotes $1 \times 1$ convolution, $\theta$ is a process parameter, and $F_{res}^{Contour}$ denotes the result obtained after residual link. Finally, we matrix dot product $F'_{hw}$ with $F_{res}^{Contour}$ to get the final output $F_{pca}$, the above process can be expressed as:

$$F_{pca} = F'_{hw} \odot F_{res}^{Contour} \tag{13}$$

**Up-sampling.** Finally, we up-sample the FAA and PCA processed features and the clothing-related features by a factor of 8 in order to facilitate the calculation of the loss with the ground truth. The up-sampling process can be expressed as:

$$F_{faa}^{up} = Up(F_{faa}) \tag{14}$$

$$F_{cloth}^{up} = Up(F_{cloth}) \tag{15}$$

$$F_{pca}^{up} = Up(F_{pca}) \tag{16}$$

Where $Up(\cdot)$ denotes the up-sampling function, $F_{faa}^{up}$, $F_{cloth}^{up}$, $F_{pca}^{up}$ are clothing-unrelated, clothing-related, and pedestrian contour features obtained after up-sampling.

### 3.3 Loss Function

In the clothing classifier, we introduce Clothes-based Adversarial Loss [8] as a loss function and use a two-stage training strategy. In the first stage, we train the clothing classifiers to distinguish clothes, and in the second stage, we fix the parameters of the classifiers and force the backbone to learn clothing-unrelated features to improve the ability of the classifiers to discriminate between clothing-related features and clothing-unrelated features. We use $G_\vartheta$ and $Clo_p(\cdot)$ to denote the backbone and the clothes classifier, respectively, where $\vartheta$ and $p$ are the parameters of the backbone and classifier, respectively. Given sample $x_i$, its corresponding clothing label is denoted as $y_i^C$ and the identity label is denoted as $y_i^{ID}$. We pass the output features of the backbone to the clothes classifier after L2 regularization, and the following equation can represent the loss function:

$$\mathcal{L}_C = -\sum_{i=1}^{N} \log \frac{e^{\left(G_{l2} \cdot \frac{p_{y_i^C}}{\tau}\right)}}{\sum_{j=1}^{N_c} e^{\left(G_{l2} \cdot \frac{p_j}{\tau}\right)}} \tag{17}$$

Where $G_{l2}$ denotes the result after L2 regularisation of $G_\vartheta(x_i)$, $p_j$ denotes the weight of the $j$-th clothing classifier, $N$ is the size of batch size, $N_c$ is the number of garments in the training set, and $\tau \in R^+$ is a temperature parameter. In the second stage of training, we use $\mathcal{L}_{CA}$ as the loss function:

$$\mathcal{L}_{CA} = -\sum_{i=1}^{N} \sum_{c=1}^{N_C} q(c) \log \frac{e^{\left(G_{l2} \cdot \frac{p_c}{\tau}\right)}}{e^{\left(G_{l2} \cdot \frac{p_c}{\tau}\right)} + \sum_{j \in S_i^-} e^{\left(G_{l2} \cdot \frac{p_j}{\tau}\right)}} \tag{18}$$

In the ID classifier, our goal is to classify people with different identities and, in the inference phase, only use the clothing-unrelated features extracted by the backbone for final evaluation, thus achieving controlled disentanglement. Given instance $x_i$ its with identity label $P_i$. the loss function is computed as follows:

$$\mathcal{L}_{ID} = -\sum_{i=1}^{N} \log\left(\frac{y(x_i, P_i)}{\sum_{j=1}^{N_{ID}} y(x_i, P_j)}\right) \tag{19}$$

$N$ and $N_{ID}$ are the number of instances and body points, respectively, and $y(x_i, P_i)$ denotes the probability of predicting $x_i$ as $P_i$. In the human reconstruction branch, we use the following loss function:

$$\mathcal{L}_R = \frac{1}{N}\sum_{i=1}^{N} l_1(Y_i, T_i) \tag{20}$$

Where $Y$ is a ternary representing the clothing-related features, clothing-unrelated features, and pedestrian contour extracted from the backbone, respectively, and $T$ is also a ternary representing the ground truth of the clothing-related features, clothing-unrelated features, and pedestrian contour.

## 4 Experiments

### 4.1 Datasets and Evaluation Protocols

We evaluate our proposed model on three mainstream datasets: LTCC-ReID [19], PRCC-ReID [20], and Vc-Clothes [21], respectively. LTCC is a dataset with variable environments and frequent changes in pedestrian clothing, containing 17,119 labeled images with 152 IDs and 478 clothing sets. Compared to the LTCC, the PRCC dataset has more images totaling 33,698, of which there are 221 IDs, three different camera views, pedestrians wearing the same clothes in shots one and two, and different clothes in shots one and three. The Vc-Clothes dataset is a set of synthetic cloth-changing data generated through the GTA5 game engine. The dataset contains 19,060 images captured from 4 cameras covering 512 fictional identities. Each identity is presented wearing one to three different sets of clothes. For model evaluation, we use rank-k (R@K) and mean average precision (mAP). We perform the same comparison for the PRCC, LTCC, and Vc-Clothes datasets in both Same-Clothes (SC) and Cloth-Changing (CC) scenarios, respectively. Where SC indicates that the people in the query and gallery images are wearing the same clothes, and CC indicates they are wearing different clothes.

### 4.2 Implementation Details

We use ResNet-50 [23] after pre-training in ImageNet1k [22] as the backbone network, and we remove the last downsampling of ResNet-50 to better extract deep features. For data enhancement, we use random cropping, inversion, and erasure, and the final image is resized for output at a size of 384×192. We train the network on NVIDIA RTX3090 GPUs, and for the LTCC, PRCC, and Vc-Clothes datasets, the batch size is set to 64, i.e., each batch contained 8 instances of 8 people with different identities. The learning rate is initialized to $3.5 \times 10^{-4}$ for the LTCC and PRCC datasets during training, and the learning rate is reduced by 90% per 20 epochs compared to the initial learning rate. Our training unfolds in two phases; in the first phase, we optimize for the $\mathcal{L}_C + \mathcal{L}_{ID} +$

Contribution Title (shortened if too long)       9

$\mathcal{L}_R$ loss function, and in the second phase, we optimize for the sum of all loss functions $\mathcal{L}_C + \mathcal{L}_{ID} + \mathcal{L}_R + \mathcal{L}_{CA}$.

**Table 1.** Comparison with the state of-the art methods on PRCC and LTCC dataset

| Method | Venue | PRCC-ReID | | | | LTCC-ReID | | | |
|---|---|---|---|---|---|---|---|---|---|
| | | Standard | | Cloth-Changing | | Standard | | Cloth-Changing | |
| | | R@1 | mAP | R@1 | mAP | R@1 | mAP | R@1 | mAP |
| HACNN[24] | CVPR 18 | 82.5 | - | 21.8 | - | 60.2 | 26.7 | 21.6 | 9.3 |
| PCB[25] | ECCV 18 | 99.8 | 97 | 41.8 | 38.7 | 65.1 | 30.6 | 23.5 | 10 |
| OSNet[26] | ICCV 19 | - | - | - | - | 67.9 | 32.1 | 23.9 | 10.8 |
| IANet[27] | CVPR 19 | 99.4 | 98.3 | 46.3 | 46.9 | 63.7 | 31 | 25 | 12.6 |
| GI-ReID[32] | CVPR 22 | 86 | - | 33.3 | - | 63.2 | 29.4 | 23.7 | 10.4 |
| UCAD[33] | IJCAI 22 | 96.5 | - | 45.3 | - | 74.4 | 34.8 | 32.5 | 15.1 |
| CAL[34] | CVPR 22 | 100 | 99.8 | 55.2 | 55.8 | 74.2 | 40.8 | 40.1 | 18 |
| DLAW[35] | TIP 23 | 98.42 | 98.7 | 56.2 | 57.1 | - | - | - | - |
| AIM[36] | CVPR 23 | 100 | 99.9 | 57.9 | 58.3 | 76.3 | 41.1 | 40.9 | 19.1 |
| SCNet[37] | ACM MM | 100 | 97.8 | 61.3 | 59.9 | 76.3 | 43.6 | 47.5 | 25.5 |
| CCFA[38] | CVPR 23 | 99.6 | 98.7 | 61.2 | 58.4 | 75.8 | 42.5 | 45.3 | 22.1 |
| IRM[39] | CVPR 24 | - | - | 54.2 | 52.3 | 75.8 | 52 | - | - |
| VersReID[40] | TPAMI 24 | - | - | 60.7 | 61.4 | - | - | - | - |
| **Ours** | | **100** | **99.9** | **65.4** | **63.3** | **77.9** | **45.1** | **50.9** | **29.8** |

**Table 2.** Comparison with the state of the art methods on Vc-Clothes dataset

| Methods | Venue | Vc-Clothes | | | |
|---|---|---|---|---|---|
| | | Same-Clothes | | Cloth-Changing | |
| | | Rank@1 | mAP | Rank@1 | mAP |
| HACNN[24] | CVPR 18 | 68.6 | 69.7 | 49.6 | 50.1 |
| MGN[41] | ACM MM 18 | - | - | - | - |
| PCB[25] | ECCV 18 | 87.7 | 74.6 | 62.0 | 62.2 |
| ABD-Net[42] | ICCV 19 | - | - | - | - |
| OSNet[26] | ICCV 19 | - | - | - | - |
| ISP[43] | CVPR 19 | 94.5 | 94.7 | 72.0 | 72.1 |
| FSAM[44] | CVPR 21 | 94.7 | 94.8 | 78.6 | 78.9 |
| CAL[34] | CVPR 22 | 92.9 | 87.2 | 81.4 | 81.7 |
| GI-ReID[32] | CVPR 22 | - | - | 64.5 | 57.8 |
| SCNet[37] | ACM MM 23 | 94.9 | 89.6 | 90.1 | 84.4 |
| IRM[39] | CVPR 24 | - | - | 90.1 | 80.1 |
| **Ours** | | **97.5** | **96.4** | **93.7** | **88.5** |

### 4.3    Comparison with the State-Of-The-Art Methods

We first compare the model in this paper with a variety of state-of-the-art ReID methods on PRCC and LTCC datasets, including HACNN [24], PCB [25], OSNet [26], IANet [27], UCAD [30], CAL [31], etc., where HACNN [24], PCB [25], OSNet [26 ], IANet [27] are normal ReID methods and the rest are cloth-changing ReID methods. As shown in Table 1, our proposed method achieves advanced performance on both PRCC and LTCC datasets. In the case of cloth-changing, our proposed method achieves 65.4% Rank-1 and 63.3% mAP on the PRCC dataset, which is an improvement of 4.1% in Rank-1 and 1.9% in mAP compared to the existing state-of-the-art methods. Our proposed method also achieves cutting-edge performance on the LTCC dataset, with our Rank-1 and mAP of 77.9% and 45.1% without cloth-changing compared to the state-of-the-art method by 1.6% and 1.5%, respectively. In the case of cloth-changing, our Rank-1 and mAP are 50.9% and 29.8%, respectively, which is a 3.4% and 4.3% improvement over the state-of-the-art method.

Table 3. Ablation experiments of the reconstruction method.

| Baseline | Unrelated | Related | Contour | LTCC-ReID Cloth-Changing | | Vc-Clothes Cloth-Changing | |
|---|---|---|---|---|---|---|---|
| | | | | Rank@1 | mAP | Rank@1 | mAP |
| √ | | | | 40.2 | 20.1 | 83.7 | 82.9 |
| √ | √ | | | 44.9 | 23.5 | 86.6 | 85.3 |
| √ | √ | √ | | 47.6 | 27.3 | 89.4 | 87.2 |
| √ | √ | √ | √ | **50.9** | **29.8** | **93.9** | **88.5** |

We also conducted experiments on a larger dataset, Vc-Clothes, and the results are shown in Table 2. Our model achieves 93.7% Rank-1 and 88.5% mAP on the Vc-Clothes dataset in the cloth-changing scenario. Compared with the state-of-the-art methods, our method improves Rank-1 and mAP by 3.6% and 4.1%, respectively. The experimental results in Tables 1 and 2 demonstrate the effectiveness of our proposed method, proving that controlled feature reconstruction can effectively improve the accuracy of cloth-changing person re-identification.

### 4.4 Ablation Study

**Ablation Experiments of the Different Reconstruction Methods.** In order to investigate the effectiveness of different reconstruction methods, we perform ablation experiments on the reconstruction methods, as shown in Table 3. When no reconstruction is performed, Rank-1 and mAP are 40.2% and 20.1% on the LTCC dataset, while Rank-1 and mAP are 83.7% and 82.9% on the Vc-Clothes dataset. When reconstructing clothing-unrelated features, Rank-1 and mAP are improved by 4.7% and 3.4% on the LTCC dataset and 2.9% and 2.4% on the Vc-Clothes dataset, respectively, compared to the baseline, which proves that reconstructing clothing-unrelated features is effective in improving the accuracy of person re-identification. When both clothing-unrelated and clothing-related features are reconstructed, Rank-1 and mAP improve by 7.4% and 7.2% on the LTCC dataset and 5.7% and 4.3% on the Vc-Clothes dataset, respectively, compared to the baseline. This shows that reconstructing both clothing-unrelated and clothing-related features at the same time enables the model to learn clothing-unrelated features better. When reconstructing clothing-unrelated features, clothing-related features, and pedestrian contour simultaneously, Rank-1 improves by 10.7% and 10.2%, and mAP improves by 9.7% and 5.6% on both datasets compared to baseline, respectively. By controllably reconstructing clothing-unrelated features, clothing-related features, and pedestrian contour, the accuracy of CC-ReID can be significantly improved.

**Ablation Experiments within the Feature Reconstruction Branch.** We also conduct ablation experiments on the LTCC and Vc-Clothes datasets for the far away attention (FAA) mechanism, person contour attention (PCA) mechanism in the feature reconstruction branch, as shown in Table 4. The first row of Table 4 indicates the accuracy of reconstruction directly using clothing-unrelated features map, clothing-related features map, and pedestrian contour map, while the second row indicates the use of far away attention mechanism on top of feature reconstruction, and the third row
indicates the use of both far away attention mechanism and person contour attention mechanism.

*Far Away Attention (FAA) Mechanism.* We perform ablation experiments by replacing the FAA to demonstrate its effectiveness. As shown in row 2 of Table 4, when using only the far away attention mechanism, the model achieves a Rank-1 of 3.3% and a mAP improvement of 3.1% on the LTCC dataset compare to the baseline. The model also achieve a rank-1 of 3.8% and a mAP improvement of 2.2% on the Vc-Clothes dataset. The far away attention mechanism enables the global modeling of clothing-unrelated features by computing the interaction between any two positions.

**Table 4.** Ablation experiments within the feature reconstruction branch.

| FAA | PCA | LTCC-ReID CC | | Vc-Clothes CC | |
|---|---|---|---|---|---|
| | | Rank@1 | mAP | Rank@1 | mAP |
| | | 42.3 | 22.4 | 86.7 | 84.5 |
| √ | | 45.6 | 25.5 | 90.5 | 86.7 |
| √ | √ | **50.9** | **29.8** | **93.9** | **88.5** |

*Person Contour Attention (PCA) Mechanism.* Row 3 of Table 4 shows person contour attention ablation experiments on the LTCC and Vc-Clothes datasets. When using the person contour attention mechanism on top of the far away attention mechanism, the model improves Rank-1 by 5.3% and mAP by 4.3% on the LTCC dataset. Also, Rank-1 and mAP on the Vc-Clothes dataset are improved by 3.4% and 1.8%, respectively. The person contour attention mechanism improves the contrast between the pedestrian contour and the background by aggregating and enhancing different levels of edge information, making the pedestrian contour clearer.

**Ablation Experiments with different attention mechanisms.** We also conducte ablation experiments using SE[45], Channel[48], Spatial[46], and CBAM[47] attention to replace far away and person contour attention mechanisms to demonstrate the effectiveness of our proposed method. The experimental results are shown in Fig. 4, where the Rank-1 of the model on the LTCC and Vc-Clothes datasets are 46.8% and 88.9% when using SE attention [45]. This metric is 46.1% and 88.5% when using spatial attention [46], 44.8% and 86.9% when using CBAM attention [47], and 45.4% and 87.6% when using channel attention [48]. All the results mentioned above are lower than the 50.9% and 93.9% achieved by the far away and person contour attention mechanism. The above experimental results show that using the far away and person contour attention mechanisms to process clothing-unrelated features and pedestrian contour maps outperforms traditional attention mechanisms.

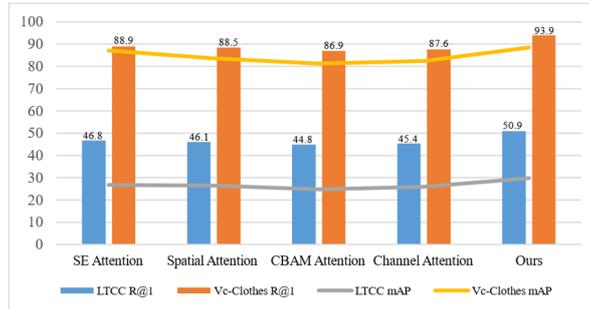

**Fig. 4.** Ablation experiments use SE, Channel, Spatial, and CBAM to replace far away and person contour attention mechanisms.

### 4.5  Visual Results

As shown in Fig. 5, we visualize the baseline method and FRD-ReID using GradCAM [63]. On the PRCC dataset, FRD-ReID pays more attention to face, arm, and leg features than the baseline method, suggesting that FRD-ReID can pay more attention to clothing-unrelated features through controlled reconstruction of the human reconstruction branch (HRB). With HRB, the features learned by FRD-ReID can highlight the head region and describe the hand and leg regions more clearly. As shown in Fig. 6, we also visualize the detection results of the model, where green borders indicate correct retrieval and red borders indicate incorrect retrieval. The visualization results show that FRD-ReID achieves more accurate identification than the baseline method when significant clothing style and color changes. Therefore, controllable reconstruction of backbone extracted features can make the model pay more attention to clothing-unrelated features and improve re-identification accuracy in clothing-changing scenarios.

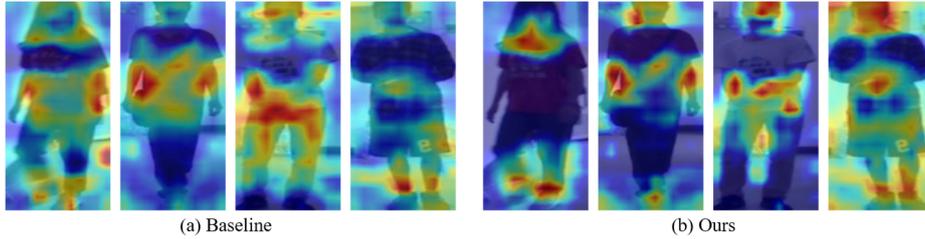

**Fig. 5.** Visualization of baseline and the FRD-ReID on the PRCC dataset.

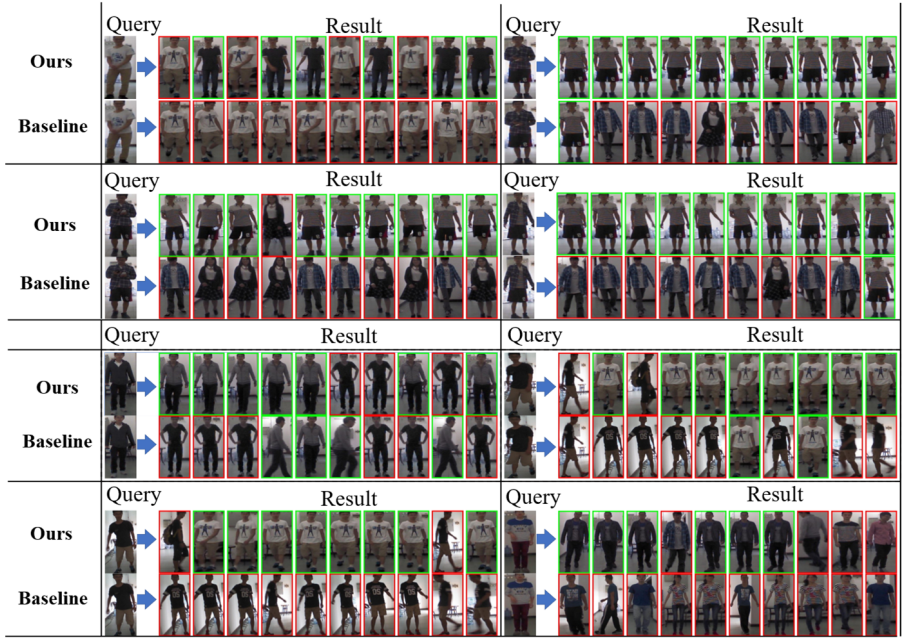

**Fig. 6.** The inference results of the FRD-ReID and baseline on PRCC.

## 5   Conclusion

In this paper, we propose a novel network named features reconstruction disentanglement ReID (FRD-ReID) for cloth-changing person re-identification. We innovatively introduce the human reconstruction branch (HRB) to disentangle clothing-unrelated features from clothing-related features in a controlled manner. In HRB, for clothing-unrelated features, we propose the far away attention mechanism to globally model sparsely distributed features. For pedestrian contour maps, we propose the person contour attention mechanism, which aggregates and enhances different levels of edge information. Finally, extensive experiments are conducted on the PRCC, LTCC, and Vc-Clothes datasets to demonstrate the effectiveness of our proposed method.

14      F. Author and S. Author14. Hong, Peixian, et al. "Fine-grained shape-appearance mutual learning for cloth-changing person re-identification." Proceedings of the IEEE/CVF conference on computer vision and pattern recognition. 2021.
15. Chen J, Jiang X, Wang F, et al. Learning 3D shape feature for texture-insensitive person re-identification[C]//Proceedings of the IEEE/CVF Conference on Computer Vision and Pattern Recognition. 2021: 8146-8155.
16. Xu, Wanlu, et al. "Adversarial Feature Disentanglement for Long-Term Person Re-identification." IJCAI. 2021.
17. Li, Yu-Jhe, et al. "Learning shape representations for clothing variations in person re-identification." arXiv preprint arXiv:2003.07340 (2020).
18. Guo, Peini, et al. "Semantic-aware Consistency Network for Cloth-changing Person Re-Identification." Proceedings of the 31st ACM International Conference on Multimedia. 2023.
19. Qian, Xuelin, et al. "Long-term cloth-changing person re-identification." Proceedings of the Asian Conference on Computer Vision. 2020.
20. Yang, Qize, Ancong Wu, and Wei-Shi Zheng. "Person re-identification by contour sketch under moderate clothing change." IEEE transactions on pattern analysis and machine intelligence 43.6 (2019): 2029-2046.
21. Wan, Fangbin, et al. "When person re-identification meets changing clothes." Proceedings of the IEEE/CVF Conference on Computer Vision and Pattern Recognition Workshops. 2020.
22. Deng, Jia, et al. "Imagenet: A large-scale hierarchical image database." 2009 IEEE conference on computer vision and pattern recognition. Ieee, 2009.
23. He, Kaiming, et al. "Deep residual learning for image recognition." Proceedings of the IEEE conference on computer vision and pattern recognition. 2016.
24. Li, Wei, Xiatian Zhu, and Shaogang Gong. "Harmonious attention network for person re-identification." Proceedings of the IEEE conference on computer vision and pattern recognition. 2018.
25. Sun, Yifan, et al. "Beyond part models: Person retrieval with refined part pooling (and a strong convolutional baseline)." Proceedings of the European conference on computer vision (ECCV). 2018.
26. Zhou, Kaiyang, et al. "Omni-scale feature learning for person re-identification." Proceedings of the IEEE/CVF international conference on computer vision. 2019.
27. Hou, Ruibing, et al. "Interaction-and-aggregation network for person re-identification." Proceedings of the IEEE/CVF conference on computer vision and pattern recognition. 2019.
28. Fu, Yang, et al. "Horizontal pyramid matching for person re-identification." Proceedings of the AAAI conference on artificial intelligence. Vol. 33. No. 01. 2019.
29. Zhu, Kuan, et al. "Identity-guided human semantic parsing for person re-identification." Computer Vision–ECCV 2020: 16th European Conference, Glasgow, UK, August 23–28, 2020, Proceedings, Part III 16. Springer International Publishing, 2020.
30. Yan, Y. M., et al. "Weakening the influence of clothing: universal clothing attri- bute disentanglement for person re-identification." Proceedings of the 31st International Joint Conference on Artificial Intelligence. Vienna, Austria: Morgan Kaufmann. 2022.
31. Gu, Xinqian, et al. "Clothes-changing person re-identification with rgb modality only." Proceedings of the IEEE/CVF Conference on Computer Vision and Pattern Recognition. 2022.
32. Jin, Xin, et al. "Cloth-changing person re-identification from a single image with gait prediction and regularization." Proceedings of the IEEE/CVF Conference on Computer Vision and Pattern Recognition. 2022.